\documentclass{article}

\PassOptionsToPackage{numbers,sort&compress}{natbib}
\usepackage[main, final]{neurips_2026}

\usepackage[utf8]{inputenc}
\usepackage[T1]{fontenc}
\usepackage{hyperref}
\usepackage{url}
\usepackage{booktabs}
\usepackage{amsfonts}
\usepackage{amsmath}
\usepackage{amssymb}
\usepackage{nicefrac}
\usepackage{microtype}
\usepackage{xcolor}
\usepackage{graphicx}
\usepackage{multirow}
\usepackage{enumitem}
\usepackage{algorithm}
\usepackage{algpseudocode}
\usepackage{booktabs}
\usepackage{float}
\usepackage{float}
\usepackage{booktabs}
\usepackage{tabularx}
\usepackage{array}

\usepackage{caption}
\title{Clinical Graph-JEPA: Predictive Patient-State Knowledge Graphs for Cognitive Decision Support}

\author{}

\author{
Kushagra Yadav$^{1}$,
Nalin Prabhath$^{2}$,
Amit Lamba$^{2}$,
James E. Schrager$^{2}$, \\
\textbf{Goeun Han}$^{2}$,
\textbf{Yining Mao}$^{2}$ \\
$^{1}$New York University,
$^{2}$University of Chicago \\
\texttt{ky2684@nyu.edu} \\
\texttt{\{nalinprabhath, hangoeun, ynmao\}@uchicago.edu} \\
\texttt{\{lamba, fjschrag\}@chicagobooth.edu}
}
\begin{document}

\maketitle

\begin{abstract}
% Clinical text contains rich evidence about patient state, but converting it into reliable, structured knowledge graphs remains difficult because extraction errors, ontology mismatch, missing relations, and temporal ambiguity propagate into downstream decision support systems. We propose a framework for clinical knowledge graph construction that combines multi-agent extraction, ontology-aware normalization, critic-based verification, and Graph-JEPA-based latent refinement. The central idea is to treat the clinical knowledge graph as a predictive patient-state representation rather than a static extraction artifact. Given clinical notes or transcripts, the system constructs an initial graph, verifies extracted entities and relations against clinical constraints, and learns to predict masked or missing graph regions from observed context. This enables the model to capture latent dependencies among symptoms, medications, behaviors, labs, and diagnoses. We evaluate the framework on clinical text datasets using extraction quality, graph completion, downstream prediction, and retrieval-augmented reasoning metrics.
Clinical records contain rich evidence about patient state, but converting that evidence into reliable, structured knowledge graphs remains difficult because extraction errors, ontology mismatch, missing relations, and temporal ambiguity can propagate into downstream systems. We propose a clinical knowledge graph construction and refinement framework that combines multi-agent relation proposal, ontology-aware normalization, deterministic evidence scoring, and JEPA-based latent refinement. Rather than treating a clinical knowledge graph as a static extraction artifact, we treat it as a predictive patient-state representation. For each admission, the system constructs an evidence-scored graph from structured MIMIC-IV records and inferred clinical cross-links, then learns to recover held-out clinical relations from the observed graph context. We evaluate the refiner with leakage-free leave-one-out edge recovery (MRR and Hits@k) and held-out batch-mask evaluation (AUC and MRR). To isolate the contribution of discharge-note context, we compare a note-embedding-free configuration with a note-augmented configuration that injects real discharge-note representations only into note-grounded entities. Under the same cohort and evaluation protocol, entity-grounded note injection improves overall leave-one-out MRR by 31\% relative improvement.

% Because real discharge notes are available for only a subset of admissions, we report a note-free configuration that applies across the cohort and a note-augmented configuration that injects real discharge-note context only onto note-grounded entities. Entity-grounded note injection improves overall leave-one-out MRR by 52.9\% on admissions with notes.

\end{abstract}

\section{Introduction}
Structured clinical data (diagnosis, medication, and procedure tables) records \emph{what} happened to a patient but not the clinician's \emph{reasoning}: that a $\beta$-blocker was started \emph{for} the patient's atrial fibrillation, that a presenting chest pain \emph{indicated} the acute coronary syndrome that was ultimately worked up. That reasoning lives in the free-text note. A large language model (LLM) can extract it as graph edges, but those inferred edges are exactly the ones with no table-level provenance, so a drafted clinical KG is a mixture of certain structural facts and uncertain inferred relations of widely varying quality. Accordingly, the initial LLM-generated clinical knowledge graph should be treated as an evidence-scored draft rather than a final representation of patient state. 

% We improved missing critical relationships(INDICATES, CONFIRMS) recovery by 34-60\%.

% We take the position that the right way to obtain a reliable clinical knowledge graph is not a one-shot extractor but a \emph{refiner}: a model that, given an already-drafted, evidence-scored graph, predicts which edges belong. We instantiate this as a \textbf{ClinG-JEPA world model} --- a self-supervised graph encoder trained by masked-latent prediction, with a lightweight readout that recovers held-out edges. Two findings drive the design. First, notes are \emph{partial}: in a longitudinal MIMIC-IV \cite{johnson2023mimiciv} cohort only $64.4\%$ of admissions carry a discharge note, so a deployable system cannot assume one is present. We therefore evaluate two refiner configurations that share the same consolidated evidence-scored graph --- a \emph{note-free} model (Option~A) that uses no real discharge-note features, and a \emph{note-augmented} model (Option~B) that uses the available real discharge notes. Second, \emph{how} the note is injected matters more than \emph{whether} it is: a diffuse, admission-global note vector slightly hurts, whereas localising the same vector onto the entities the note grounds nearly doubles inferred-edge recovery \cite{hamilton2017graphsage}.

We take the position that the right way to obtain a reliable clinical knowledge graph is not a one-shot extractor but a \emph{refiner}: a model that, given an already-drafted, evidence-scored graph, predicts which edges belong. We instantiate this as a \textbf{Clinical Graph-JEPA world model} --- a self-supervised graph encoder trained by masked-latent prediction, with a lightweight readout that recovers held-out edges. The model supports two input configurations. Option A is \emph{note-free} and relies only on the consolidated evidence-scored graph, while Option B augments the same graph with Clinical-ModernBERT representations localized to the entities grounded by the discharge note. This distinction preserves a note-independent deployment path while allowing us to quantify the contribution of note context. The results further show that injection strategy is important: a diffuse admission-level note vector lands at the no-note level, whereas localizing the same representation to note-grounded entities substantially improves inferred-edge recovery.

% a diffuse admission-level note

% Two findings drive the design. First, notes are \emph{partial}: in a longitudinal MIMIC-IV \cite{johnson2023mimiciv} cohort only $64.4\%$ of admissions carry a discharge note, so a deployable system cannot assume one is present. We therefore evaluate two refiner configurations that share the same consolidated evidence-scored graph --- a \emph{note-free} model (Option~A) that uses no real discharge-note features, and a \emph{note-augmented} model (Option~B) that uses the available real discharge notes. Second, \emph{how} the note is injected matters more than \emph{whether} it is: a diffuse, admission-global note vector slightly hurts, whereas localising the same vector onto the entities the note grounds nearly doubles inferred-edge recovery \cite{hamilton2017graphsage}.

\paragraph{Contributions.}
(1) A consolidated, evidence-scored clinical-graph construction that fuses a deterministic MIMIC-IV backbone with LLM-inferred relations and a 14-signal per-edge score. (2) A Clinical Graph-JEPA world model that refines these drafts, evaluated by a leakage-free leave-one-out edge-recovery protocol. (3) Two deployment options --- \textbf{Option~A} (note-free) and \textbf{Option~B} (note-augmented). (4) The empirical finding that real discharge-note information is most useful when injected \emph{locally} into the entities it grounds, improving Option B over Option A from $0.364$ to $0.477$ overall leave-one-out MRR.

\section{Related Work}
% \paragraph{Clinical KG extraction.} Multi-agent and decomposed LLM pipelines extract entities and relations from clinical text, typically separating an entity-extraction stage from a relation stage; the ACI-Bench golden reference we validate against was built this way. We depart from this line by treating extraction as a \emph{draft} to be refined rather than a final artifact, and by learning the refiner self-supervised.

\textbf{Clinical KG extraction.} Prior clinical KG systems commonly decompose graph construction into entity identification, entity normalization, relation extraction, and schema or evidence-based validation \cite{himm2017Hetionet}. Recent LLM-based and multi-agent approaches further specialize these stages to improve coverage and reduce confusion between entity recognition and relation reasoning. However, these methods generally treat the extracted graph as the final representation consumed by downstream systems. Such a graph may combine directly observed structural facts with uncertain text-inferred relations that are incomplete, weakly supported, or incorrectly linked. Our work separates graph construction from graph refinement: the initial graph is treated as an \emph{evidence-scored draft}, and a self-supervised Graph-JEPA refiner learns to recover plausible held-out relations from the surrounding graph context.

\textbf{Self-supervised and world models.} Joint-embedding predictive architectures (JEPA) \cite{lecun2022path, assran2023selfsupervisedlearningimagesjointembedding} and BYOL-style latent prediction learn representations by predicting masked latents under an exponential-moving-average (EMA) target with stop-gradient, avoiding contrastive collapse without negatives \cite{grill2020bootstraplatentnewapproach}. We adapt this objective to graphs as the pre-training stage of our refiner, yielding a single shared encoder we treat as a world model over clinical state.

\textbf{Knowledge-graph completion.} Bilinear scoring functions such as DistMult \cite{yang2015embeddingentitiesrelationslearning} provide an efficient edge-recovery readout. We apply one over the \emph{frozen} world-model encoder, trained with an InfoNCE \cite{oord2019representationlearningcontrastivepredictive} objective and type-matched negative sampling, so that recovery quality reflects the pre-trained representation rather than a jointly-tuned decoder.

\textbf{Clinical language models.} Domain-pretrained bidirectional encoders (e.g.\ ModernBERT \cite{warner2024smarterbetterfasterlonger} and its clinical variant, Clinical-ModernBERT \cite{lee2025clinicalmodernbertefficientlong}) provide note representations. Our contribution is not the note encoder but \emph{where} its output is injected into the graph.

\section{Method: Clinical Graphs as Predictive World Models}
We formulate clinical knowledge graph refinement as a world-modeling problem over
structured patient-state memory. The observation is a structured MIMIC-IV \cite{johnson2023mimiciv}
admission record together with its source-faithful clinical narrative, the state
is a typed clinical knowledge graph, and revision actions correspond to
retaining, reviewing, pruning, or proposing relations. The graph acts as an
externalized patient-state representation that organizes diagnoses,
medications, procedures, microbiology, and admission context into an
auditable structure. Our system does not make autonomous clinical decisions; it
improves the structured representation used by downstream retrieval and
decision-support systems.

\begin{figure*}[t]
    \centering
    \includegraphics[
        width=0.92\textwidth,
        height=0.48\textheight,
        keepaspectratio
    ]{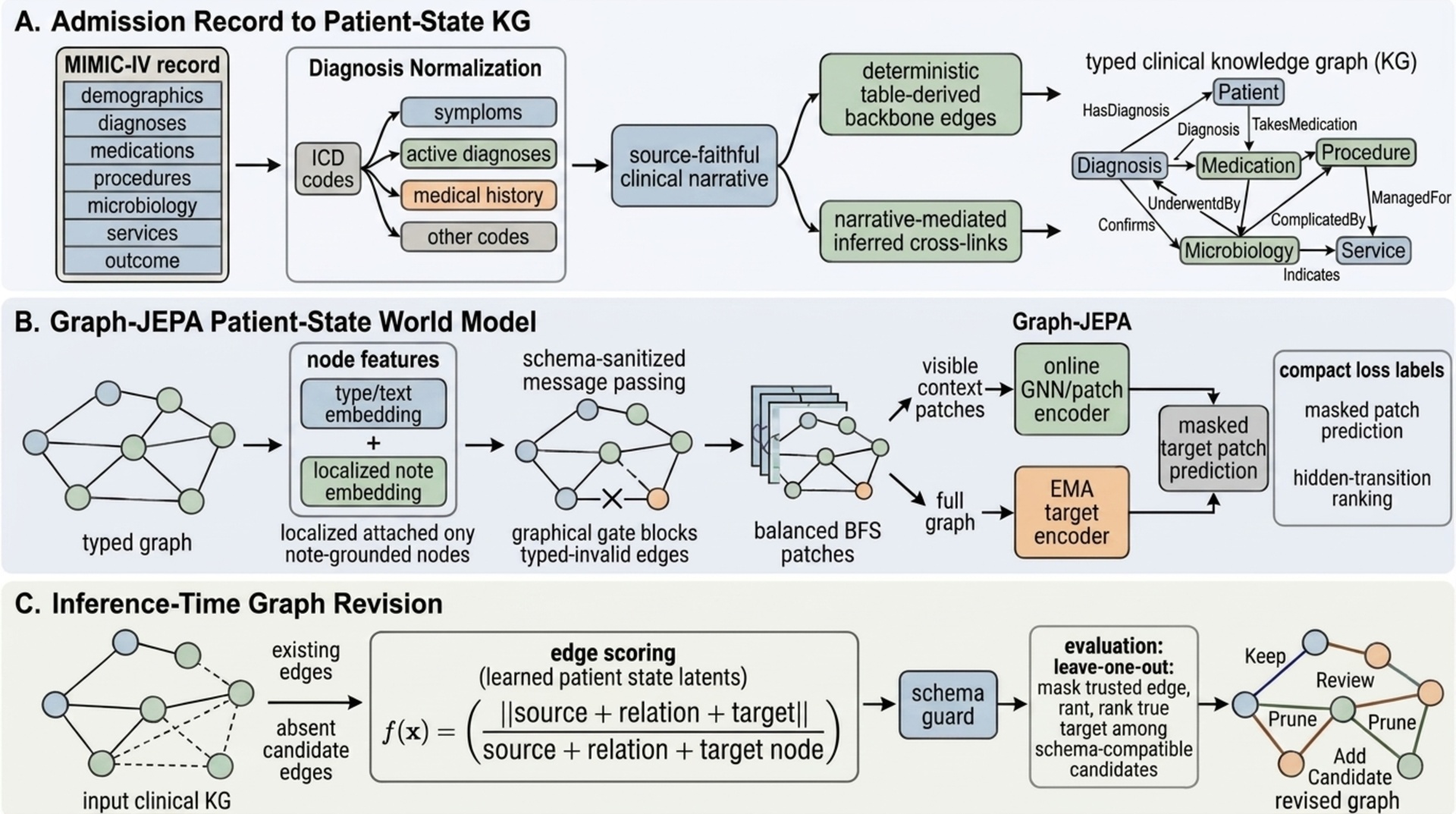}
    \caption{
    Overview of Clinical Graph-JEPA.
    (A) Staged extraction converts MIMIC-IV record into evidence-grounded patient-state knowledge graphs.
    (B) Graph-JEPA learns representations through schema-aware message passing and masked BFS-patch prediction.
    (C) Inference combines learned edge scoring with schema validation for graph revision, while LOO evaluation measures trusted-edge recovery.
    }
    \label{fig:clinical-graph-jepa}
\end{figure*}

Figure~\ref{fig:clinical-graph-jepa} summarizes the pipeline. Panel A shows how
structured MIMIC-IV records are converted into patient-state graphs, combining
deterministic table-derived edges with narrative-mediated inferred clinical
cross-links. Panel B shows Graph-JEPA training, where node features combine
entity text embeddings with localized note context, followed by masked patch
pretraining and schema-aware graph-revision fine-tuning. Panel C shows
inference-time revision, where learned relation plausibility and patch-level
predictive consistency are combined before a schema guard assigns the final
revision action.

\subsection{Consolidated clinical graph}
\label{sec:graph}

Each admission becomes one \emph{consolidated knowledge graph} fusing two
evidence streams. A \emph{deterministic backbone} is read from the MIMIC-IV
structured tables, patient $\rightarrow$ diagnoses, medications, procedures,
microbiology, and services
(\texttt{HAS\_DIAGNOSIS}, \texttt{TAKES\_MEDICATION},
\texttt{UNDERWENT\_PROCEDURE}, etc) each at confidence $1.0$. Over this scaffold,
four specialist agents propose the inferred relations the tables do not encode:
\texttt{MANAGED\_FOR} (medication$\rightarrow$diagnosis),
\texttt{CONFIRMS} (procedure$\rightarrow$diagnosis),
\texttt{COMPLICATED\_BY} (diagnosis$\rightarrow$diagnosis), and
\texttt{INDICATES} (presenting symptom$\rightarrow$diagnosis).

\textbf{Deterministic evidence scoring.}
Every inferred edge receives a deterministic support vector computed from model
confidence, biomedical knowledge bases, ontology proximity, and note
provenance. These signals include RxNorm linkage, RxNav/MED-RT may-treat
evidence, Hetionet \cite{himm2017Hetionet} treatment and disease-similarity relations, OMOP concept
similarity and ancestor distance, and note-grounding checks
(\texttt{prov\_in\_note}, \texttt{prov\_ratio}). The same evidence-scored graph
is used for both deployment options; the only difference between options is
whether note features are provided to the refiner.

\subsection{Clinical Graph-JEPA world model}
\label{sec:wm}

We adapt the JEPA principle to typed clinical knowledge graphs using a masked-node objective, rather than directly following the original Graph-JEPA subgraph-prediction pipeline \cite{skenderi2025graphlevelrepresentationlearningjointembedding}. The refiner trains in two phases. \emph{Phase~1} is self-supervised
masked-latent prediction: a context encoder embeds a masked graph and predicts
the latents assigned by an EMA target encoder to the hidden state. The encoder
is a typed graph transformer (TransformerConv \cite{shi2021maskedlabelpredictionunified}) over the consolidated graph. Node inputs include a
type embedding, a hashed-entity embedding, projected numeric features, and, only
under Option~B, a localized note vector. \emph{Phase~2} freezes the encoder and
trains an edge-recovery readout with an InfoNCE objective using $K{=}8$
type-matched negatives. This trains the world model to recover graph-state
relations from the surrounding patient-state memory rather than from labels or
clinical outcomes. Additional details on the entity-localized encoder, masked-node JEPA objective, frozen edge recovery, and leave-one-out evaluation are provided in Appendix~\ref{app:gnn-patching}--\ref{app:candidate-generation}.

\subsection{Deployment Options A and B}
\label{sec:options}

We evaluate two configurations to isolate the contribution of localized discharge-note representations. 

\paragraph{Option A --- note-free graph refiner.}
Option A consumes the consolidated clinical graph without receiving the discharge note or its embedding as a node feature. Its inputs include node type and text features, demographic attributes, typed relations, schema information, edge confidence, and deterministic edge-support metadata. Schema-aware message passing prevents typed-invalid edges from influencing node representations. Graph-JEPA is pretrained by masking connected BFS patches and predicting their latent representations from the visible graph context using an online graph encoder and an EMA target encoder. During revision training, schema-valid trusted edges are treated as positive graph-state transitions, while type-compatible corruptions, typed-invalid observed edges, invalid reversed triples, and weak low-confidence edges provide negatives. A hidden-edge ranking objective removes trusted edges from message passing and trains the model to rank the true target above hard schema-compatible alternatives. Because it requires no
note at inference time, Option~A applies uniformly to all admissions.

\paragraph{Option B --- note-augmented refiner.}
Option~B keeps the same graph, schema, encoder, training losses, and evaluation
protocol as Option~A, but augments node features when a real clinician-authored discharge note is available. The
note is encoded once using Clinical-ModernBERT
(\texttt{Simonlee711/Clinical\_ModernBERT}; 768-dimensional mean-pooled
embedding) and injected locally onto note-grounded entities. Nodes not grounded
in the note receive a zero note vector. This entity-grounded placement is essential.  A single admission-global note vector degrades recovery because it is diffuse and largely redundant with the graph structure extracted from the note. Localizing the same note representation
onto the entities it discusses instead provides targeted context for relation
revision.

%  A single admission-global note
% vector degrades recovery because it is diffuse and largely redundant with the
% graph structure extracted from the note. Localizing the same note representation
% onto the entities it discusses instead provides targeted context for relation
% revision.

% Thus, Option~A is the fully deployable note-free refiner, while Option~B asks
% whether localized note context improves graph revision on the subset where notes
% exist. The two options differ only in the note feature.

\section{Results}
\subsection{Data and Cohort Construction}
\label{sec:data}
% We use MIMIC-IV, a de-identified critical-care EHR \cite{johnson2023mimiciv}. From the full set of $546{,}028$ admissions we identify $175{,}165$ \emph{eligible} patients --- those with at least one admission meeting completeness criteria ($\geq 3$ diagnoses, $\geq 3$ medications, $\geq 1$ service) --- and draw a seed-fixed sample of $1{,}000$ patients, retaining \emph{all} of each patient's admissions to preserve longitudinal structure. This yields $2{,}969$ admissions (mean $3.0$ per patient), and the resulting preprocessed corpus contains $4{,}000$ patient-admission graph records. Critically, a discharge note is present for only $1{,}912/2{,}969$ ($64.4\%$) of these admissions; $1{,}057$ ($35.6\%$) have none. At the patient level, $523$ patients are fully noted, $251$ have no noted admission, and the remainder are partial. This partiality is the empirical basis for the two-option design of Section~\ref{sec:options}.

We use a fixed random seed sample of MIMIC-IV, a de-identified critical-care EHR \cite{johnson2023mimiciv}, comprising $3{,}018$ unique patients, producing $4{,}000$ admission-level patient graphs. The cohort includes patients with single and multiple admissions, with a maximum of 7 admissions per patient. Where a discharge note is available, the graph is paired with it and its $768$-dimensional Clinical-ModernBERT embedding. We evaluate two configurations (Section~\ref{sec:options}) on the same seed-fixed $80/10/10$ graph-level split of $3{,}200$ training, 400 validation, and 400 test graphs.  This controlled design isolates the contribution of localized discharge-note embeddings under an otherwise identical training and evaluation protocol.

\subsection{Protocol}
\label{sec:protocol}
We evaluate by \emph{leave-one-out} (LOO) edge recovery, the fair link-prediction protocol for a refiner: a single edge is removed (with its inverse, to prevent leakage), the rest of the consolidated graph is kept as context, and the model must recover the held-out edge by ranking the true target against same-type candidates under a \emph{filtered} protocol (other true tails of the query excluded). Ranking is deterministic and reproducible. We report mean reciprocal rank (MRR) and Hits@$k$ overall and per relation, with chance computed from candidate-set size. We additionally report a held-out batch-mask AUC and a \emph{non-obvious} AUC that excludes trivial patient-hub targets. The evaluation is explicitly of a \emph{refiner}: it improves an already-constructed draft and presupposes the consolidated graph as context, not a cold-start predictor.

\subsection{Internal results (MIMIC-IV held-out): Option A vs Option B}
\label{sec:internal}
Table~\ref{tab:loo} is the central result. \textbf{Option~A} (note-free) and \textbf{Option~B} (note-augmented, on noted admissions) are measured on identical data, split, seed, \emph{and training recipe} ($4{,}000$ graphs; $3{,}200/400/400$ split; $n{=}8{,}283$ LOO edges); they differ only in the note feature, so $\Delta$ is the pure effect of the note.

\begin{table}[h]
\centering
\begin{tabular}{lccc}
\toprule
Inferred relation & Option~A (note-free) & Option~B (note-augmented) & $\Delta$ \\
\midrule
\texttt{MANAGED\_FOR}     & 0.326 ± 0.048 & 0.423 ± 0.029 & $+0.097$ \\
\texttt{INDICATES}        & 0.469 ± 0.051 & 0.637 ± 0.047 & $+0.168$ \\
\texttt{COMPLICATED\_BY}  & 0.405 ± 0.056 & 0.527 ± 0.032 & $+0.122$ \\
\texttt{CONFIRMS}         & 0.395 ± 0.035 & 0.506 ± 0.055 & $+0.111$ \\
\midrule
\textbf{Overall LOO MRR}  & \textbf{0.364 ± 0.045} & \textbf{0.477 ± 0.026} & $\mathbf{+0.113}$ \\
\bottomrule
\end{tabular}
\caption{Leave-one-out edge recovery (MRR), Option~A vs Option~B, on the four inferred relations. Identical recipe; only the note feature differs.}
\label{tab:loo}
\end{table}

Option~B improves \emph{every} inferred relation, with the largest gains on \texttt{INDICATES} (the presenting-symptom$\rightarrow$diagnosis link the note narrates most directly, $0.469\!\rightarrow\!0.637$) and \texttt{COMPLICATED\_BY} ($0.405\!\rightarrow\!0.527$, ${\approx}1.3\times$). The smallest gain is on \texttt{MANAGED\_FOR} ($+0.097$), where Option~A already recovers comparatively well ($0.326$). 
Option A remains capable of recovering held-out edges without note features, although it consistently underperforms the note-augmented Option B.
Crucially, Option~A stays well above chance on every relation, which makes it a viable standalone configuration and the fallback inside Option~B for note-less admissions.

\begin{table}[h]
\centering
\begin{tabular}{lcc}
\toprule
Metric & Option~A & Option~B \\
\midrule
Batch-mask AUC            & \textbf{0.811 ± 0.003} & 0.805 ± 0.013 \\
Non-obvious AUC           & 0.712 ± 0.031 & \textbf{0.800 ± 0.009} \\
Batch-mask MRR            & 0.358 ± 0.009 & \textbf{0.375 ± 0.005} \\
LOO MRR (overall)         & 0.364 ± 0.045 & \textbf{0.477 ± 0.026} \\
LOO Hits@1                & 0.182 ± 0.046 & \textbf{0.303 ± 0.033} \\
LOO Hits@10               & 0.821 ±0.037 & \textbf{0.917 ± 0.011} \\
\bottomrule
\end{tabular}
\caption{Global metrics. The note (Option~B) helps on every refiner-relevant measure; the hub-saturated batch-mask AUC is not discriminating.}
\label{tab:global}
\end{table}

Table~\ref{tab:global} reports global metrics. Option~B raises overall LOO Hits@1 ($0.182\!\rightarrow\!0.303$) and Hits@10 ($0.821\!\rightarrow\!0.917$) and, on held-out batch-mask scoring, the discriminating \emph{non-obvious} AUC ($0.712\!\rightarrow\!0.80$). The undifferentiated batch-mask AUC is saturated by trivial patient-hub edges and is not the meaningful measure; the non-obvious AUC and the LOO MRR are. The global \texttt{NOTE}-node configuration was evaluated across multiple seeded experiments.

\begin{table}[h]
\centering
\begin{tabular}{lc}
\toprule
Configuration & Overall LOO MRR \\
\midrule
Option~A: note-free (no note)            & 0.364 ± 0.045  \\ 
Global \texttt{NOTE} node (diffuse note) & 0.352 ± 0.020 \\
\textbf{Option~B: entity-grounded note} & \textbf{0.477 ± 0.026 } \\
\bottomrule
\end{tabular}
\caption{Note-placement ablation. The same note vector helps only when localised onto the entities it grounds.}
\label{tab:placement}
\end{table}

\subsection{Ablation: note placement}
Isolating the note feature on identical data confirms that where the note is placed is decisive (Table~\ref{tab:placement}).
A diffuse global note lands at the no-note level ($0.352$ vs.\ $0.364$, within noise), while entity-grounding delivers the gain ($0.477$) --- where the note is placed is what matters. More further ablations can be found in Appendix~\ref{app:aci-evaluation}--\ref{sec:graph-qa-ablation}

% Isolating the note feature on identical data confirms that \emph{where} the note is placed is decisive (Table~\ref{tab:placement}). Global note placement is less effective than entity-grounded placement; localising the same vector onto note-grounded entities aligns narrative evidence with graph structure and is the form used by Option~B. More further ablations can be found in Appendix~\ref{app:aci-evaluation}--\ref{sec:graph-qa-ablation}

\subsection{Where does the recovery come from? A context analysis}
To attribute Option~B's recovery we isolate, per inferred relation, three context conditions: a \emph{floor} (only the deterministic backbone present), a \emph{cascade} (the backbone plus earlier inferred relations), and the full LOO \emph{ceiling} (all other edges present). Table~\ref{tab:cascade} shows that the deterministic backbone alone already supplies most of the recoverable context (e.g.\ \texttt{INDICATES} $0.491$ at the floor vs.\ $0.637$ at the ceiling), with the remaining headroom coming from the other inferred cross-links. This indicates the model exploits the free, certain structure first and the inferred relations second --- consistent with the refiner framing.

\begin{table}[h]
\centering
\begin{tabular}{lccc}
\toprule
Inferred relation & Floor (backbone) & Cascade & Ceiling (LOO) \\
\midrule
\texttt{MANAGED\_FOR}    & 0.300 $\pm$ 0.007 & 0.300 $\pm$ 0.007 & 0.423 $\pm$ 0.029 \\
\texttt{CONFIRMS}        & 0.349 $\pm$ 0.017 & 0.386 $\pm$ 0.029 & 0.506 $\pm$ 0.055 \\
\texttt{COMPLICATED\_BY} & 0.374 $\pm$ 0.018 & 0.433 $\pm$ 0.009 & 0.527 $\pm$ 0.032 \\
\texttt{INDICATES}       & 0.491 $\pm$ 0.012 & 0.527 $\pm$ 0.015 & 0.637 $\pm$ 0.047 \\
\bottomrule
\end{tabular}
\caption{Context analysis for Option~B (MRR). The deterministic backbone (floor) provides most of the context; inferred cross-links add the remainder.}
\label{tab:cascade}
\end{table}

\subsection{Limitations}
\label{sec:limitations}
This study has three main limitations. First, all graphs are derived from MIMIC-IV and reflect a single hospital system, so the findings may not generalize to other institutions, populations, coding practices, or clinical workflows. Second, the leave-one-out evaluation measures recovery of held-out relations within an evidence-scored draft graph rather than correctness against an independently adjudicated clinical graph. Finally, the study does not evaluate prospective clinical utility, causal validity, or patient outcomes. The model should therefore be viewed as a graph-refinement component for research and decision-support workflows, not as an autonomous clinical decision-maker.

\section{Discussion}
% \textbf{Why entity-grounding works.} The note and the graph carry overlapping information --- the entities were extracted \emph{from} the note --- so a global note summary is largely redundant and adds variance. Placing the note on the specific entities it grounds turns it into a localised prior that distinguishes note-discussed entities from purely structural ones, which is exactly the distinction the inferred relations depend on. This is consistent with the negative result we observed: feeding the per-edge evidence scores directly into the encoder also failed to help. Diffuse, graph-level features add little to a structure-first world model; localised, edge- or entity-specific signal is what helps.

\textbf{Why entity-grounding works.} The note and the graph carry overlapping information—the entities were extracted from the note. A global note summary broadcasts the same context across the graph, whereas entity grounding preserves which entities that context actually describes. Placing the note on the specific entities it grounds turns it into a localised prior that distinguishes note-discussed entities from purely structural ones, which is exactly the distinction the inferred relations depend on. This is consistent with two further results we observed: feeding the
per-edge evidence scores directly into the encoder also failed to help, and a separate global NOTE
node doesn't improve at the no note level. Diffuse, graph-level features add little to a structure-first world model; localised, edge- or entity-specific signal is what helps.

\textbf{Why two options.} Reporting a single note-dependent model would overstate deployable accuracy,
since a third of admissions have no note. Option A provides a note-embedding-free baseline, while Option B measures the additional value of localized note context under otherwise identical conditions. This framing supports deployment in settings with or without note access while providing a controlled estimate of the note contribution.

% Reporting a single note-dependent model would overstate deployable accuracy, since a third of admissions have no note. Option~A establishes the universally-available floor; Option~B is the achievable ceiling where notes exist. The two-option framing makes the system honest about coverage while still capturing the note's value.

\section{Conclusion}
Clinical Graph-JEPA treats an LLM-drafted clinical knowledge graph as an evidence-scored representation to be \emph{refined} rather than accepted as final. The refiner learns to recover held-out, schema-valid clinical relations from the surrounding patient-graph context. We compare two configurations on the same 4,000-graph cohort: \textbf{Option A} refines the graph without direct discharge-note embeddings, whereas \textbf{Option B} injects localized discharge-note representations into note-grounded entities.Under a matched training and evaluation protocol, entity-grounded note features improve Option B over Option A from $0.364$ to $0.477$ overall leave-one-out MRR on the evaluation subset.

% Because real discharge notes are available for only a subset of admissions, we evaluate two configurations: \textbf{Option A}, which refines the graph without real-discharge-note features and provides coverage across the cohort, and \textbf{Option B}, which adds localized real-discharge-note features when available.

Future work will evaluate the refiner against independently clinician-annotated relations and across external institutions and patient populations. Additional directions include enforcing patient-disjoint and temporally separated evaluation, developing calibrated graded edge-revision policies, evaluating downstream retrieval and decision-support utility.

% Use ack only in final camera-ready version. It is hidden in anonymized mode by the NeurIPS style.
% \begin{ack}
% Add funding, competing interests, and acknowledgments here only for the final version.
% \end{ack}

\bibliographystyle{plainnat}
\bibliography{references}

\appendix
\section{Additional Method Details}
\label{app:method-details}

\subsection{MIMIC Graph Schema}
\label{app:schema}

For MIMIC-derived graphs, each admission graph is represented as
$G_i=(V_i,E_i)$ with typed nodes and directed typed edges. The node type set is
\begin{equation}
\begin{aligned}
\mathcal{T}_{\mathrm{MIMIC}}=\{&
\mathrm{PATIENT},\ \mathrm{DIAGNOSIS},\ \mathrm{MEDICATION},\\
&\mathrm{PROCEDURE},\ \mathrm{MICROBIOLOGY},\ \mathrm{SERVICE}\}.
\end{aligned}
\end{equation}
When available, the schema can also include laboratory-test nodes, but in the
MIMIC-IV extract used here laboratory values are not available and laboratory
features are therefore not used as measured result nodes.

The relation set contains deterministic backbone relations and inferred clinical
cross-links:
\begin{equation}
\begin{aligned}
\mathcal{R}_{\mathrm{MIMIC}}=\{&
\mathrm{HAS\_DIAGNOSIS},\
\mathrm{TAKES\_MEDICATION},\
\mathrm{UNDERWENT\_PROCEDURE},\\
&\mathrm{HAD\_MICROBIOLOGY},\
\mathrm{MANAGED\_BY\_SERVICE},\
\mathrm{TARGETS\_ORGANISM},\\
&\mathrm{MANAGED\_FOR},\
\mathrm{CONFIRMS},\
\mathrm{COMPLICATED\_BY},\
\mathrm{INDICATES}\}.
\end{aligned}
\end{equation}

The deterministic backbone relations connect the patient node to observed
entities in the structured record. Examples include
\[
\mathrm{PATIENT}
\xrightarrow{\mathrm{HAS\_DIAGNOSIS}}
\mathrm{DIAGNOSIS},
\]
\[
\mathrm{PATIENT}
\xrightarrow{\mathrm{TAKES\_MEDICATION}}
\mathrm{MEDICATION},
\]
and
\[
\mathrm{PATIENT}
\xrightarrow{\mathrm{UNDERWENT\_PROCEDURE}}
\mathrm{PROCEDURE}.
\]
Microbiology susceptibility data provide direct antibiotic--organism evidence,
represented as
\[
\mathrm{MEDICATION}
\xrightarrow{\mathrm{TARGETS\_ORGANISM}}
\mathrm{MICROBIOLOGY}.
\]

The inferred cross-link schema includes clinically meaningful relations not
stored directly as MIMIC relational columns:
\[
\mathrm{MEDICATION}
\xrightarrow{\mathrm{MANAGED\_FOR}}
\mathrm{DIAGNOSIS},
\]
\[
\mathrm{PROCEDURE}
\xrightarrow{\mathrm{CONFIRMS}}
\mathrm{DIAGNOSIS},
\]
\[
\mathrm{DIAGNOSIS}
\xrightarrow{\mathrm{COMPLICATED\_BY}}
\mathrm{DIAGNOSIS},
\]
and
\[
\mathrm{DIAGNOSIS}
\xrightarrow{\mathrm{INDICATES}}
\mathrm{DIAGNOSIS}.
\]
Presenting symptoms are represented as diagnosis-layer nodes with a presenting
attribute, so symptom--diagnosis links use the same node type while preserving
the symptom role in node metadata.

\subsection{MIMIC Record Normalization}
\label{app:mimic-normalization}

Each graph is built from a flattened MIMIC-IV admission record. The record
contains patient and admission identifiers, demographics, admission and discharge
metadata, diagnoses, medications, procedures, microbiology, services, and
outcome fields. Diagnoses are partitioned into four clinically distinct buckets:
presenting symptoms, active diagnoses, medical history, and other/external
codes. This partition prevents historical or external-status codes from being
treated as active admission diagnoses.

Medications retain dose, unit, route, and count metadata when available.
Microbiology records retain specimen, organism, antibiotic, and susceptibility
interpretation. Services and transfer-derived location information are retained
as admission context. These fields provide the source of graph nodes and the
metadata attached to nodes and edges.

\subsection{Graph Construction}
\label{app:extraction-pipeline}

Given a normalized admission record $r_i$, graph construction first creates a
deterministic node set
\begin{equation}
V_i =
V_i^{\mathrm{patient}}
\cup V_i^{\mathrm{dx}}
\cup V_i^{\mathrm{med}}
\cup V_i^{\mathrm{proc}}
\cup V_i^{\mathrm{micro}}
\cup V_i^{\mathrm{service}} .
\end{equation}
The deterministic backbone edge set is
\begin{equation}
E_i^{\mathrm{backbone}}
=
B(r_i),
\end{equation}
where $B$ adds direct table-derived relations such as diagnoses, medications,
procedures, microbiology, services, and susceptibility-supported
antibiotic--organism links.

To infer clinical cross-links, the structured record is rendered into a
source-faithful synthetic note narrative $n_i=N(r_i)$. A source-faithful synthetic narrative is generated from the structured record so the extraction agents can propose the inferred cross-links. We use two types of notes across the pipeline: synthetic note generated from structured records and real discharge notes authored by clinicians. The narrative is constrained to mention only facts present in $r_i$. Relation-specific extraction agents then operate over candidate pairs of existing graph nodes:
\begin{equation}
E_i^{\mathrm{cross}}
=
A_{\mathrm{managed}}(V_i,n_i)
\cup
A_{\mathrm{confirms}}(V_i,n_i)
\cup
A_{\mathrm{complicated}}(V_i,n_i)
\cup
A_{\mathrm{indicates}}(V_i,n_i).
\end{equation}
The final graph is
\begin{equation}
G_i =
\left(
V_i,\,
E_i^{\mathrm{backbone}}\cup E_i^{\mathrm{cross}}
\right).
\end{equation}

The extraction agents are constrained to link only nodes already present in
$V_i$. Candidate edges with unmatched endpoints, invalid type signatures,
self-loops, duplicate entries, or unsupported evidence are removed during
validation.

\subsection{Provenance and Edge Support}
\label{app:validation}

Each edge stores its source node, target node, relation type, confidence, source
field or evidence text, and provenance metadata. Backbone edges are directly
grounded in MIMIC tables and are assigned full confidence. Inferred cross-links
store cited evidence from the generated narrative and are further annotated with
deterministic support signals.

For medication--diagnosis relations, support includes biomedical semantic
similarity and treatment evidence from resources such as RxNorm/RxNav and
Hetionet \cite{himm2017Hetionet}. Additional support features include OMOP concept similarity or
ontology proximity when available, as well as provenance checks indicating
whether the cited evidence appears in the narrative. These scores are retained
as continuous edge attributes rather than collapsed into hard-coded rules. This
allows downstream models to learn how to use support signals while preserving
the original high-recall graph.

\subsection{Graph Unit and Entity Identity}
\label{app:entity-resolution}

The training unit is an admission-level graph. Entities are grounded in the
structured MIMIC record before relation extraction, and inferred edges are
restricted to those existing entities. Therefore, the current MIMIC pipeline does
not require corpus-level embedding-based entity resolution to construct the
training graphs. Node normalization is performed within the admission record,
and node metadata retains the original MIMIC-derived names, codes, and source
fields needed for provenance.

\subsection{Node and Note Features}
\label{app:node-note-features}

Each node is represented by a base embedding of its type and canonical text:
\begin{equation}
b_v=f_\phi(\tau(v),\mathrm{text}(v)).
\end{equation}
When an admission-level note embedding $n_i$ is available, it is localized to
nodes grounded in the note. Let $a_v\in\{0,1\}$ indicate whether node $v$ is
grounded. The final node feature is
\begin{equation}
x_v=[b_v,\ a_v n_i].
\end{equation}
If no note embedding is available, the note component is zero.

By default, a node is considered grounded when it participates in an edge whose
provenance check indicates support in the note. We also support grounding by
surface-name match to the note, or assigning the note embedding to all
non-patient nodes. This keeps note context localized to clinically grounded
entities rather than broadcasting it uniformly to the whole graph.

\subsection{GNN Encoder, Graph Patching, and Target Encoding}
\label{app:gnn-patching}

The Graph-JEPA encoder uses typed message passing over the schema-sanitized
message graph. Relation labels are embedded and used as edge attributes in the
GNN. The final node latent is
\begin{equation}
z_v=W_o h_v^{(L)} .
\end{equation}

The graph is partitioned into $K$ local patches using balanced multi-source BFS.
Patch representations are mean-pooled node latents:
\begin{equation}
q_j=\frac{1}{|P_j|}\sum_{v\in P_j} z_v .
\end{equation}
Patch positional features include relative patch size, patch degree, and
random-walk return features on the coarsened patch graph.

For each graph, context patches $C$ and target patches $T$ are sampled. The
online encoder receives visible context patches and learned mask content for
hidden patches. The EMA target encoder receives the full graph and is updated by
\begin{equation}
\bar{\theta}\leftarrow m\bar{\theta}+(1-m)\theta .
\end{equation}
For each $j\in T$, the predictor maps the context patch state and positional
features to $\hat q_j$.

\subsection{Hyperparameters}
\label{app:hyperparameters}
\begin{table}[h]
\centering
\caption{Principal hyperparameters used in the pipeline}
\label{tab:v16_hyperparameters}
\small
\begin{tabular}{lll}
\toprule
\textbf{Stage} & \textbf{Hyperparameter} & \textbf{Value} \\
\midrule

\multirow{5}{*}{Graph encoder}
& Architecture & \texttt{TransformerConv} \\
& Hidden dimension & 128 \\
& Number of layers & 2 \\
& Attention heads & 4 \\
& Relation embedding dimension & 32 \\
\midrule

\multirow{6}{*}{JEPA pretraining}
& Pretraining epochs & 60 \\
& Batch size & 16 \\
& Learning rate & $1\times10^{-3}$ \\
& Target-node mask ratio & 0.40 \\
& EMA coefficient & $0.996 \rightarrow 0.9999$ \\
& Prediction loss & Cosine distance \\
\midrule

\multirow{7}{*}{Edge readout}
& Readout epochs & 40 \\
& Encoder during readout & Frozen \\
& Decoder & MLP \\
& Objective & InfoNCE \\
& Negative samples per positive & 8 \\
& Edge-mask ratio & $\mathcal{U}(0.10,0.60)$ \\
& Target-relation weight & 3.0 \\
\midrule

\bottomrule
\end{tabular}
\end{table}

\subsection{Masked-Node JEPA Loss}
\label{app:revision-supervision}

The self-supervised objective predicts normalized target-encoder latents
for masked nodes. Let $\hat z_v$ be the predictor output and $\bar z_v$ the
stop-gradient target latent. The loss is cosine prediction error:
\begin{equation}
\mathcal{L}_{\mathrm{JEPA}}
=
\frac{1}{|M|}
\sum_{v\in M}
\left(
2 - 2\,
\frac{\hat z_v^\top \bar z_v}
{\|\hat z_v\|_2\|\bar z_v\|_2}
\right).
\end{equation}

\subsection{Frozen Edge Recovery}
\label{app:candidate-ranking}

After JEPA pretraining, the graph encoder is frozen and a MLP readout is
trained for edge recovery. For a candidate triple $e=(u,r,v)$, the score is
\begin{equation}
s_{\phi}(u,r,v)
=
\sum_{\ell=1}^{h}
a_{\ell}\,
\sigma\!\left(
\mathbf{w}^{u}_{\ell}\cdot\mathbf{Z}_{u}
+
\mathbf{w}^{v}_{\ell}\cdot\mathbf{Z}_{v}
+
\mathbf{w}^{r}_{\ell}\cdot\mathbf{Z}_{r}
+
b_{\ell}
\right)
+
b_{0},
\end{equation}

where $\mathbf{Z}_{u}$ and $\mathbf{Z}_{v}$ denote the source and destination
node embeddings, respectively, and $\mathbf{Z}_{r}$ denotes the relation
embedding. The vectors $\mathbf{w}^{u}_{\ell}$,
$\mathbf{w}^{v}_{\ell}$, and $\mathbf{w}^{r}_{\ell}$ are learned weights for
the $\ell$-th hidden unit, $b_{\ell}$ is its bias, $a_{\ell}$ is the learned
output-layer weight, $b_{0}$ is the output bias, $h$ is the number of hidden
units, and $\sigma(\cdot)$ is the ReLU activation function.
For each hidden positive edge, negative targets are sampled from nodes of the
same type. With candidate set $\mathcal{C}_i=\{e_i\}\cup\mathcal{N}_i$, the
ranking loss is
\begin{equation}
\mathcal{L}_{\mathrm{rank}}
=
-\frac{1}{|H|}
\sum_{i\in H}
\log
\frac{\exp(s_\phi(e_i)/T)}
{\sum_{e\in\mathcal{C}_i}\exp(s_\phi(e)/T)} .
\end{equation}

% For leave-one-out recovery, one trusted edge is removed from the message graph.
% The source and relation are held fixed, and the true target is ranked against
% same-type candidate targets using the frozen encoder and MLPYeah this is right. Can you give me the whole latex subsection that I should write on this in my research paper? I think the section should start with "What exactly are we doing on the ACI bench?" Basically we are testing our evaluation on the ACI bench.  score.
% Performance is reported with MRR and Hits@$k$. This evaluates edge recovery from
% learned patient-state latents, not inference-time graph revision.

% For leave-one-out recovery, one trusted edge is removed from the message graph. The source and
% relation are held fixed, and the true target is ranked against same-type candidate targets using the
% frozen encoder and MLP.  Performance is reported with MRR and Hits@k. This evaluates edge recovery from learned patient-state latents, not inference-time graph revision

\subsection{Leave-One-Out Evaluation}
\label{sec:loo-evaluation}

For leave-one-out edge recovery, one trusted edge
$(u,r,v)$ is removed from the message-passing graph while the remaining
graph context is retained. The source node $u$ and relation type $r$ are
held fixed, and the model ranks the true target node $v$ against candidate
target nodes of the same semantic type. Other known true targets for the
same source--relation query are filtered from the candidate set.

The graph encoder and the MLP readout are kept fixed
during evaluation. Performance is reported using mean reciprocal rank
(MRR) and Hits@$k$, with $k \in \{1,3,10\}$. This protocol measures the
recovery of held-out clinical relationships from learned patient-state
latents. It does not perform inference-time graph revision, graph
completion, or iterative edge insertion.

\subsection{Candidate Edge Interpretation}
\label{app:candidate-generation}

ClinG-JEPA scores source--relation--target triples as recovery candidates. High
scores indicate that a candidate edge is plausible under the learned
patient-state representation, but they are not treated as verified clinical
facts. Unlike the modular graph-revision pipeline, we do not assign
explicit Keep, Review, Prune, or Add-Candidate actions.

\subsection{External Validation on ACI-Bench}
\label{app:aci-evaluation}

We evaluated the model on the independent ACI-Bench dataset to assess whether
the entity-grounded note-injection finding remains valid beyond the
MIMIC-based evaluation. The evaluation used all 207 available encounters and
was performed without retraining the model on ACI-Bench.

We constructed the knowledge graphs using the same multi-agent pipeline used
above for the MIMIC data. The resulting graph records were embedded using the same Clinical-ModernBERT
embedding procedure used in the primary pipeline. We evaluated three conditions
using the same model checkpoint, graph topology, candidate-generation procedure,
and leave-one-out edge-recovery protocol. Option~A used no note information,
the global-note condition, and Option~B injected the note embedding only into entities supported
by note provenance. The results are shown in
Table~\ref{tab:aci_external_validation}.

\begin{table}[H]
\centering
\small
\renewcommand{\arraystretch}{1.15}
\setlength{\tabcolsep}{5pt}
\caption{External validation on the 207-record ACI-Bench dataset. All
conditions were evaluated using the same model checkpoint and the same 5,005
rankable leave-one-out queries.}
\label{tab:aci_external_validation}
\begin{tabularx}{\linewidth}{@{}
  >{\raggedright\arraybackslash}X
  >{\centering\arraybackslash}p{0.15\linewidth}
  >{\centering\arraybackslash}p{0.15\linewidth}
  >{\centering\arraybackslash}p{0.15\linewidth}
  >{\centering\arraybackslash}p{0.15\linewidth}
@{}}
\toprule
\textbf{Condition} &
\textbf{MRR} &
\textbf{Hits@1} &
\textbf{Hits@3} &
\textbf{Hits@10} \\
\midrule
Option A: No note
& 0.475
& 0.304
& 0.561
& 0.844 \\

Global note injection
& 0.472
& 0.301
& 0.554
& 0.855 \\

Option B: Entity-grounded note injection
& \textbf{0.481}
& \textbf{0.305}
& \textbf{0.570}
& \textbf{0.865} \\
\bottomrule
\end{tabularx}
\end{table}

As shown in Table~\ref{tab:aci_external_validation}, entity-grounded note
injection achieved the highest MRR and the highest Hits@1, Hits@3, and Hits@10
values. Compared with global note injection, entity-grounded injection
improved MRR by approximately $+0.009$, with a 95\% confidence interval of
$[0.006, 0.013]$. These results indicate that the entity-grounded injection
finding also holds on the independent ACI-Bench dataset, supporting the
benefit of localizing note information to the entities it supports rather than
applying the same note representation uniformly across the graph.

% Required packages:
% \usepackage{float}
% \usepackage{booktabs}
% \usepackage{tabularx}
% \usepackage{array}

\subsection{Discharge-Note Representation Ablation}
\label{app:discharge-note-ablation}

We compared three note representations while keeping the same
entity-grounded injection strategy: a global mean embedding, uniform pooling
of local note spans, and entity-conditioned attention over local spans. The
purpose was to determine whether preserving local note information improves
graph-edge recovery over the global mean representation.
The experiment used the same datasets and configurations described above.

\begin{table}[H]
\centering
\small
\renewcommand{\arraystretch}{1.15}
\setlength{\tabcolsep}{5pt}
\caption{Discharge-note representation ablation. Values are mean $\pm$
standard deviation across ten paired seeds. All note representations were
injected only into entities with note provenance.}
\label{tab:note_representation_ablation}
\begin{tabularx}{\linewidth}{@{}
  >{\raggedright\arraybackslash}X
  >{\centering\arraybackslash}p{0.19\linewidth}
  >{\centering\arraybackslash}p{0.19\linewidth}
  >{\centering\arraybackslash}p{0.19\linewidth}
@{}}
\toprule
\textbf{Note representation} &
\textbf{LOO MRR} &
\textbf{Hits@1} &
\textbf{Hits@10} \\
\midrule
Entity-grounded global mean
& $0.468 \pm 0.021$
& $0.291 \pm 0.026$
& $0.911 \pm 0.008$ \\

Entity-grounded uniform local spans
& $\mathbf{0.478 \pm 0.020}$
& $\mathbf{0.302 \pm 0.027}$
& $\mathbf{0.915 \pm 0.005}$ \\

Entity-conditioned attention
& $0.458 \pm 0.026$
& $0.280 \pm 0.030$
& $0.911 \pm 0.011$ \\
\bottomrule
\end{tabularx}
\end{table}

The entity-grounded global-mean condition used one 768-dimensional
Clinical ModernBERT embedding for the complete discharge note and copied it to
each entity with note provenance. The uniform local-span condition instead
represented each grounded entity using a token-count-weighted mean of its
available local note spans. The attention condition used an entity-specific
query to attend over the available spans.

As shown in Table~\ref{tab:note_representation_ablation}, uniform local-span
pooling achieved the highest numerical LOO MRR ($0.478$), compared with
$0.468$ for the entity-grounded global mean. However, its paired improvement
of $+0.010$ had a 95\% confidence interval of $[-0.013,+0.032]$, which
includes zero. Entity-conditioned attention performed worse than both
alternatives, with an LOO MRR of $0.458$ and a paired difference of $-0.020$
relative to uniform local-span pooling. 

\subsection{Note-Encoder Ablation}
\label{app:note-encoder-ablation}

We evaluated whether entity-grounded note gains depend on the quality of the
note encoder. The full graph-learning and edge-recovery pipeline was kept
fixed, including the graph, entity-grounding procedure, masking strategy,
training schedule, data split, and evaluation protocol. Only the node-level
note encoder was changed: TF--IDF embeddings were compared with the existing
Clinical-ModernBERT embeddings.

\begin{table}[H]
\centering
\caption{Entity-grounded note-encoder ablation. Values are mean $\pm$
standard deviation across seeds.}
\label{tab:note-encoder-ablation}
\small
\begin{tabular}{lcc}
\toprule
Metric & Clinical-ModernBERT & TF--IDF \\
\midrule
Batch-mask MRR & $0.37 \pm 0.004$ & $0.36 \pm 0.003$ \\
Filtered LOO MRR & $\mathbf{0.47 \pm 0.02}$ & $0.38 \pm 0.04$ \\
Filtered LOO Hits@1 & $\mathbf{0.29 \pm 0.03}$ & $0.19 \pm 0.04$ \\
Filtered LOO Hits@3 & $\mathbf{0.54 \pm 0.02}$ & $0.43 \pm 0.04$ \\
Filtered LOO Hits@10 & $\mathbf{0.91 \pm 0.01}$ & $0.84 \pm 0.02$ \\
\bottomrule
\end{tabular}
\end{table}

As shown in Table~\ref{tab:note-encoder-ablation}, TF--IDF encoder underperformed Clinical-ModernBERT on all primary LOO
metrics, indicating that entity grounding alone does not account for the full
recovery gain. The result suggests that note-encoder quality also contributes
substantially to graph-edge recovery.

\subsection{Graph-QA Ablation}
\label{sec:graph-qa-ablation}

We conducted a question-answering ablation to measure the contribution of clinical note information to graph-based reasoning. We created a fixed benchmark of 20 clinical questions and corresponding answers using the GPT-5.6 Luna model and the MIMIC dataset. The benchmark included direct graph-recall questions, clinical metadata questions, and multi-hop questions requiring the composition of multiple graph relations. The same questions, graph contexts, candidate answer sets, and gold answers were evaluated under two configurations described above.

For each question, the model ranked candidate graph entities and produced a predicted answer set. We report macro entity-level F1, question-level Hits@$k$, and strict question exact match. Question Hits@$k$ measures the fraction of questions for which every required answer group contains at least one correct entity within the top-$k$ ranked candidates. Strict question exact match requires the complete predicted answer set for every answer group to exactly match the gold answer set.

\begin{center}
\captionof{table}{Option A--B graph question-answering ablation on the fixed 20-question benchmark. Values are percentages.}
\label{tab:graph-qa-note-ablation}
\begin{tabular}{lrrr}
\toprule
Metric & Option A & Option B & $\Delta$ (B--A) \\
\midrule
Macro entity F1 & 70.6 & 74.1 & +3.6 pp \\
Question Hits@1 & 65.0 & 65.0 & +0.0 pp \\
Question Hits@3 & 65.0 & 70.0 & +5.0 pp \\
Strict question exact match & 65.0 & 65.0 & +0.0 pp \\
\bottomrule
\end{tabular}
\end{center}

As shown in Table~\ref{tab:graph-qa-note-ablation}, Across the 20-question benchmark, Option~B improved macro entity F1 by 3.6 percentage points and question-level Hits@3 by 5.0 percentage points, while strict question exact-match accuracy remained unchanged at 65.0\%. 

% If the official NeurIPS 2026 author kit requires checklist.tex, uncomment after adding it.
% \newpage
% \input{checklist.tex}

\end{document}